\documentclass[11pt,letterpaper]{article}
\usepackage{emnlp2016}
\usepackage{times}
\usepackage{latexsym}
\usepackage{graphicx}
\usepackage{booktabs}

\emnlpfinalcopy

\def\emnlppaperid{20}

\title{Generating Politically-Relevant Event Data}

\author{John Beieler\\Human Language Technology Center of Excellence\\Johns Hopkins University\\
  {\tt jbeieler1@jhu.edu}}

\date{}

\begin{document}

\maketitle

\begin{abstract}
  Automatically generated political event data is an important part of
  the social science data ecosystem. The approaches for generating this data,
  though, have remained largely the same for two decades. During this time,
  the field of computational linguistics has progressed tremendously. This paper presents
  an overview of political event data, including methods and ontologies, and a set
  of experiments to determine the applicability of deep neural networks to the 
  extraction of political events from news text.
\end{abstract}

\section{Introduction}

Automated coding of political event data, or the record of who-did-what-to-whom within the context of political actions,
has existed for roughly two decades. This type of data can prove highly useful for many types of studies. Since this type of data is inherently atomic, each observation is a record of a single event between
a source and a target, it provides a disaggregated view of political events. This means that the data can be used to examine interactions below the usual monthly or yearly
levels of aggregation. This approach can be used in a manner consistent with traditional hypothesis testing that is the norm within political science \cite{gleditsch1,gleditsch2,goldstein2}. Additionally, event
data has proven useful in forecasting models of conflict since the finer time resolution allows analysts to gain better leverage over the prediction problem than is possible
when using more highly aggregated data \cite{mine,forecasting1,forecasting2,forecasting3}.

The methods used to generate this data have remained largely the unchanged during the past two decades, namely using
parser-based methods with text dictionaries to resolve candidate noun and verb phrases to actor and event categories.
The underlying coding technologies have moved slowly in updating to reflect changes in natural language processing (NLP) technology. 
These NLP technologies have now advanced to such a level, and with accompanying open-source software implementations, 
that their inclusion in the event-data coding process comes as an obvious advancement. Given this, this paper presents the beginnings of how modern 
natural language processing approaches, such as deep neural networks, can work within the context of automatically generating political event data.
The goal is to present a new take on generating political event data, moving from parser-based methods to classifier-based models for the identification
and extraction of events.  
Additionally, this paper serves as an introduction to politically-relevant coding ontologies that offer a new application domain for natural language
processing researchers. 

\section{Political Event Ontologies}

Political event data has existed in various forms since the 1970s. Two of the original political event datasets were the World Event
Interaction Survey (WEIS) and the Conflict and Peace Data Bank (COPDAB) \cite{azar,weis}. These two datasets were eventually
replaced by the projects created by Philip Schrodt and various collaborators. In general, these projects were marked by the use of the Conflict
and Mediation Event Observations (CAMEO) coding ontology and automated, machine-coding rather than human coding \cite{cameo,cameo2}.
The CAMEO ontology is made up of 20 ``top-level'' categories that encompass actions such as ``Make Statement'' or ``Protest.'' Each of these
twenty top-level categories contains finer-grained categories in a hierarchical manner. For example, the code \texttt{14} is the top-level
code for ``Protest'' with the sub-code \texttt{142} representing a general demonstration or rally. Under the code \texttt{141} is code \texttt{1411}
which codes ``demonstrate or rally for leadership change.'' Thus, as one moves down the hierarchy of CAMEO, more fine-grained events
are encountered. All told, this hierarchical scheme contains over 200 total event classifications. This ontology has served as the basis for most of the 
modern event datasets such as the Integrated Crisis Early Warning System (ICEWS) \cite{obrien}, the Global Database of Events, Language, and Tone (GDELT)\footnote{\url{gdeltproject.org}}, and the Phoenix\footnote{\url{phoenixdata.org}} dataset.

The defining feature of the CAMEO ontology is the presence of a well-defined ontology consistent of verb phrases and noun phrases used in the
coding of actions and actors. For each of the 200+ categories of event types in CAMEO, there exists a list of verb phrases that define a given
category. Similarly, the scope of what is considered a valid actor within CAMEO is defined by the noun phrases contained in the actor dictionaries.
Thus, CAMEO is scoped entirely by the human-defined noun and verb phrases contained within underlying text dictionaries. The creation of these
dictionaries is a massively costly task in terms of human labor; to date each phrase in the dictionaries was identified, defined, and formatted
for inclusion by a human coder. 

\subsection{Lower Resolution Ontologies}
 
\begin{center}
\begin{figure}
\includegraphics[width=0.54\textwidth]{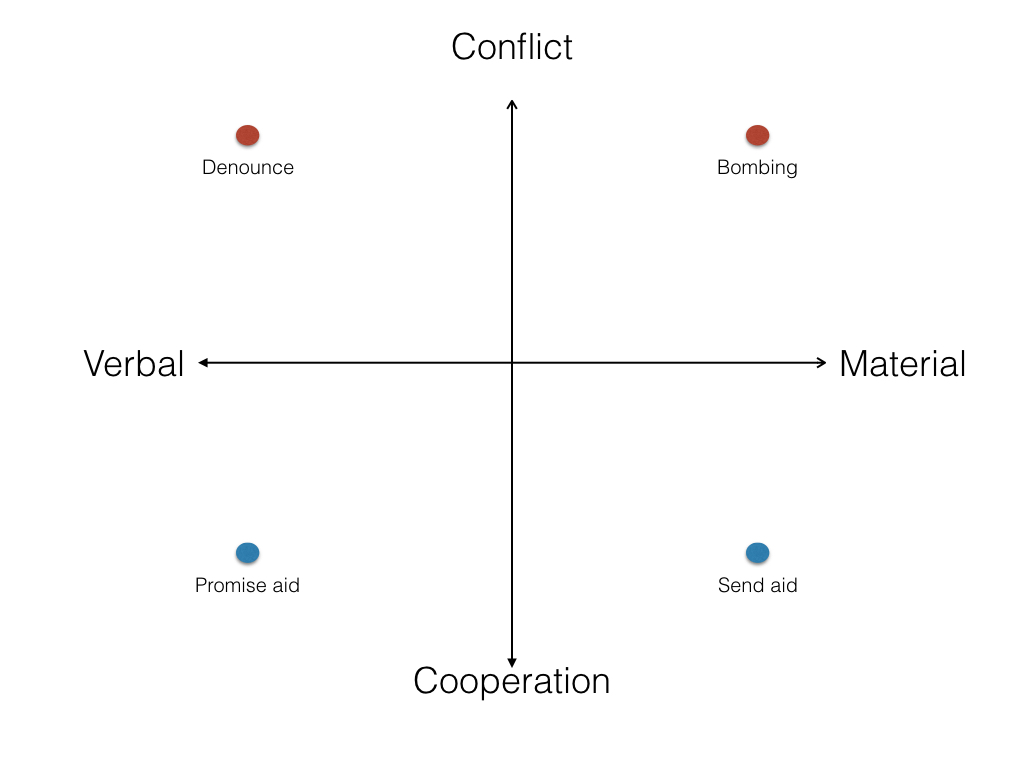}
\caption{QuadClass Event Quadrants}\label{fig:quad}
\end{figure}
\end{center}

While the CAMEO ontology offers fine-grained coding of political events within the 20 top-level categories, a small but convincing set of literature
suggests that this level of granularity is not necessary to answer many of the questions event data is used to answer. \newcite{philBalkans}, for example,
suggests that dividing the CAMEO ontology into much lower-resolution categories, known as \texttt{QuadClasses}, provides enough information to perform
accurate out-of-sample forecasts of relevant political events. Additionally, \newcite{wallach} indicates that it is possible to
recover this latent structure from coded events. These \texttt{QuadClass} variables, which are divided along conflict/cooperation
and material/verbal axises, capture the information described in the above papers. As seen in Figure \ref{fig:quad}, a given event can be placed somewhere within
the resultant quadrants based on what valence the event has (conflict or cooperation) and what realm the event occurred (verbal or material).

\noindent Since these \texttt{QuadClass} variables capture much of the information necessary, the methods discussed within this paper focus on this
rather than the full CAMEO coding ontology. 

\section{Current Approaches}

The current state-of-the-art for CAMEO-coded political event extraction is presented by the PETRARCH2\footnote{\url{https://github.com/openeventdata/petrarch2}} coder.\footnote{Other coders exist, such as BBN's ACCENT coder, but is not currently publicly available. PETRARCH2 and ACCENT approach event coding in roughly the same manner, however.} 
The main features of PETRARCH2 include a deep reliance on information from a constituency parse tree. The default parse comes from
the Stanford CoreNLP software \cite{stanford}. The exact operational details of PETRARCH2 are beyond the scope of this paper, with a complete explanation 
of the algorithm available in Norris (2016), it should suffice to say that this second version of PETRARCH makes extensive use of the actual structure of 
the parse tree to determine source-action-target event codings. This change to be more tightly coupled to the tree 
structure of the sentence allows for a clearer identification of actors and the assignment of role codes to the actors, and a more accurate 
identification of the who and whom portions of the who-did-what-to-whom equation.

At its heart, PETRARCH2 is still software to perform a lookup of terms in a set of text dictionaries. Given this, if the terms identified by the program
are incorrect then the final event coding will also be incorrect. Additionally, if the terms identified by PETRARCH2 are not in the dictionaries, but
would still be associated with a valid actor or event coding, then no event is coded. This means that no matter the algorithmic design of the event
coder, the output will remain constrained by the verb and actor dictionaries. 

The primary issue with these methods is twofold. First, the parser-based methods rely on human-created dictionaries. As noted above, this is a
labor intensive task that is not easily replicable for expansion into new ontologies; CAMEO has become the de-facto coding standard for political
events largely owing to the existence of the text dictionaries created over the course of two decades. \newcite{brendan} introduced a method that 
potentially solves the issue of developing verb patterns for the coding of political events. This approach still does not address many of the other
issues present with the current approaches to generating political event data, such as a reliance on syntactic parsers. The second issue, owing to the 
reliance on text dictionaries and parsers for the extraction of events, is the exclusively English-language nature of all available event datasets.
The next section introduces an alternative to these parser-based methods that is applicable across ontologies, is tune-able for a given problem set, and 
is capable of working cross-lingually.

\section{Statistical Approaches}

In order to replace the parser-based methods for identifying an event, a new system must
indentify to which of the four \texttt{QuadClass}
variables, Material Conflict, Material Cooperation, Verbal Conflict, or Verbal Cooperation, the event belongs. 
To accomplish this, this paper makes use of convolutional neural nets. 

This paper considers two neural architectures to classify a political event. 
The first is a 1-dimensional ConvNet with pre-trained word embedding features as described in Kim \shortcite{kim2015}. 
In short, it is a relatively shallow network with three parallel convolutional layers
 and two fully-connected layers.  The fully-connected layers contain 150 and 4 units each. 
 Dropout is applied to the two fully-connected layers so meaningful connections 
 are learned by the model. The details of the model layers are in ~Table~\ref{fig:wconv}.
 
     \begin{table}
      \begin{center}
      \begin{tabular}{cccc}
        \toprule
        Layer & Frame & Kernel & Pool \\
        \midrule
        1 & 256 & 3 & 2 \\
        2 & 256 & 4 & 2 \\
        3 & 256 & 5 & 2 \\
        \bottomrule
      \end{tabular}
      \end{center}
      \caption{Word-based convolutional layers.}\label{fig:wconv}
    \end{table}

The second model is a character ConvNet based on the work by \newcite{lecun}. The character ConvNet
is a modified implementation since multiple experiments determined the fully specification in \newcite{lecun}
underperformed other specifications. The architecture for the character ConvNet consists of 3 convolutional 
layers and 3 fully-connected layers. The convolution layers are detailed in~Table~\ref{fig:convnet}. 
The fully connected layers have 1024, 1024, and 4 output units, respectively.
    \begin{table}
      \begin{center}
      \begin{tabular}{cccc}
        \toprule
        Layer & Frame & Kernel & Pool \\
        \midrule
        1 & 256 & 7 & 3 \\
        2 & 256 & 3 & N/A \\
        3 & 256 & 3 & N/A \\
        4 & 256 & 3 & 3 \\
        \bottomrule
      \end{tabular}
      \end{center}
      \caption{Character-based convolutional layers.}\label{fig:convnet}
    \end{table}

\section{Data}

The datasets used in this paper are shown in~Table~\ref{fig:data}. Each of the ``soft-labelled'' datasets has QuadClass labels applied to them by PETRARCH2. The
use of PETRARCH2 is necessary in order to generate enough training samples for the various
classification algorithms to gain leverage over the classification task. The English corpus consists of data scraped from various online
news media sites. The Arabic
corpus labels are obtained via the use of a sentence-aligned English/Arabic corpus.\footnote{The specific 
corpus is available at \url{https://www.ldc.upenn.edu/collaborations/past-projects/gale/data/gale-pubs}.} Thus, if a sentence is labelled as \emph{Material Conflict} in the English corpus, that
label is transferred to the aligned sentence in the Arabic corpus. If multiple alignments occur the label is
transferred to each of the relevant sentences. The next dataset is the same set
of labelled Arabic sentences that were run through the machine-translation software \emph{Joshua} \cite{joshua}.
These datasets provide information for three experiments: English-language, Arabic-language, and machine translated
English. 

    \begin{table}
      \begin{center}
      \begin{tabular}{lc}
        \toprule
        Dataset & Sentences \\
        \midrule
        Soft-labelled English & 49,296  \\
        Soft-labelled Arabic  & 11,466  \\
        Machine-translated Arabic & 3,931  \\
        \bottomrule
      \end{tabular}
      \end{center}
      \caption{Data source type and size}\label{fig:data}
    \end{table}

\section{Experiments}

~Table~\ref{fig:exps} shows the results of various models for classifying a sentence
into one of four \texttt{QuadClasses}. Across the board, it is clear
that the character-based ConvNets perform much better than the
word-based models. The difference is less drastic for English-language
inputs, a 9\% difference in accuracy. For Arabic-language inputs, 
however, the difference is striking. The character model is over 20\%
more accurate than the word-based model. This is likely due to 
issues with tokenization and morphology when dealing with the
word-based models. Even more impressive is the ability of the
Arabic-language models to perform well even with a relatively small
corpus of 11,466 coded sentences. These results demonstrate that 
character-based ConvNets are appropriate and powerful models for
the classification of politically-relevant events. 

\begin{center} 
\begin{table}
\label{tab:results}
\centering
\begin{tabular}{l c c } 
\hline\hline           
   Model       &   Accuracy \\
\hline
\multicolumn{2}{c}{Word-based models} \\
\hline
English Input    &  0.85   \\
Native Arabic Input    &  0.25   \\
Machine-translated Input    &  0.60   \\
\hline
\multicolumn{2}{c}{Character-based models} \\
\hline
English input    &  0.94   \\
Arabic input     &  0.93   \\
\hline\hline
\end{tabular} 
\caption{Accuracy scores for Category Classifier}\label{fig:exps}
\end{table}
\end{center}

\vspace{-23pt}

\section{Conclusion}

This paper has demonstrated that modern approaches to natural language processing, specifically
deep neural networks, offer a promising avenue for the extraction of politically-relevant events.
The methods shown in this paper can work across both ontologies and languages offering a level
of flexibility unseen in the realm of CAMEO-coded political event data. The implementation of
these methods will allow for the exploration of languages and ontologies, as an example expanding
beyond the limits of CAMEO to code events such as crime events in Spanish-language news sources,
that will open new avenues of social science research.

While these methods are promising, there is still much work left to develop a fully operational
event extraction pipeline. In terms of classifying events, there is still the issue of handling the
many nuances of event coding. For example, if a meeting occurs between three actors that would typically
lead to nine coded events when handling the various combinations of actors and binary relations. Additionally,
the methods presented on this paper do not touch upon the extraction of actor information. This is another area
for which modern NLP approaches, such as semantic role labelling, are highly applicable and will likely improve
on the state-of-the-art.

\section*{Acknowledgments}

This work was supported by the DARPA Quantitative Crisis Response program.

\bibliography{emnlp2016}
\bibliographystyle{emnlp2016}

\end{document}